# Recognizing student identification numbers from the matric templates using a modified U-net architecture

Filip Pavičić
*course: Research Seminar (supervisor: Assoc. Prof., Ph.D. Marko Čupić)*
University of Zagreb Faculty of Electrical Engineering and Computing
Zagreb, Croatia
filip.pavicic@fer.hr

*Abstract*—This paper presents an innovative approach to student identification during exams and knowledge tests, which overcomes the limitations of the traditional personal information entry method. The proposed method employs a matrix template on the designated section of the exam, where squares containing numbers are selectively blackened. The methodology involves the development of a neural network specifically designed for recognizing students' personal identification numbers. The neural network utilizes a specially adapted U-Net architecture, trained on an extensive dataset comprising images of blackened tables. The network demonstrates proficiency in recognizing the patterns and arrangement of blackened squares, accurately interpreting the information inscribed within them. Additionally, the model exhibits high accuracy in correctly identifying entered student personal numbers and effectively detecting erroneous entries within the table. This approach offers multiple advantages. Firstly, it significantly accelerates the exam marking process by automatically extracting identifying information from the blackened tables, eliminating the need for manual entry and minimizing the potential for errors. Secondly, the method automates the identification process, thereby reducing administrative effort and expediting data processing. The introduction of this innovative identification system represents a notable advancement in the field of exams and knowledge tests, replacing the conventional manual entry of personal data with a streamlined, efficient, and accurate identification process.

## I. Introduction

In today's education system, the increasing number of students poses challenges when conducting knowledge checks and correcting exams. The process of correcting exams is often time-consuming and prone to errors due to the large volume of work involved. This article presents an innovative approach aimed at streamlining exam administration and correction processes.

The proposed approach revolves around the use of a unique student identification number to identify students during exams, replacing the traditional method of entering personal data. Instead, a method is introduced that involves blackening squares with numbers inside a matrix template on the designated part of the exam. This standardized approach eliminates reliance on individual handwriting and enables automated student identification using a neural network. Additionally, this method allows for the detection of incorrect entries in the matrix display. In the event of an incorrect entry, the examiner retains the ability to manually identify the student and rectify the error.

The implementation of a customized neural network for square recognition within the matrix template enables automatic identification and interpretation of student personal information. Leveraging deep learning models, the neural network analyzes the layout of the black squares and interprets the information based on learned patterns. As a result, this approach facilitates a fast and accurate student identification process during exams.

It is worth noting that this approach is not limited solely to the education sector. Similar concepts can be applied in other industries where users possess unique identification numbers, such as healthcare or administration. Implementing this technology can provide an intuitive and efficient identification method across various domains.

The subsequent sections of this article will delve into the practical implementation of student personal number recognition. This includes presenting the dataset utilized for training the network, describing the data annotation methodology, and providing a detailed explanation of the network architecture employed for identification number recognition. Furthermore, the article will present the results of network performance evaluations and discuss potential avenues for further system enhancement.

## II. Related work

The existing body of literature encompasses several scholarly papers that explore similar issues and topics.

The paper [1] from 2012 focuses on off-line restricted-set handwritten word recognition for student identification in a short answer question automated assessment system. The proposed system incorporates the Gaussian Grid and Modified Direction Feature Extraction Techniques, which achieve promising recognition rates (up to 99.08% for Modified Direction and up to 98.28% for Gaussian Grid).

Another paper [2] from 2014 introduces an off-line handwriting recognition system for identifying Thai student names in an automated assessment system. The system utilizes the Gaussian Grid and Modified Direction Feature extraction techniques on upper and lower contours, loops, and full word contour images. Results show encouraging recognition rates, with both techniques achieving 99.27% accuracy using artificial neural networks and support vector machine classifiers.

The paper [3] from 2015 introduces a Short Answer Assessment System (SAAS) that incorporates an off-line handwriting recognition system and novel combined features. The SAAS aims to automate the assessment of handwritten short answer questions, reduce marking time, and minimize errors in transcription. The proposed system utilizes advanced feature extraction techniques and achieves improved recognition rates. The SAAS system achieves a recognition accuracy of 95.99% using WRLGGF, outperforming other feature extraction techniques such as GGF, MDF, and WRLMDF. The proposed system successfully recognizes and marks examination papers while also identifying students from their name components.

While existing literature focuses on hand-written student identification, to the best of our knowledge, no studies have explored student identification based on recognizing unique student ID numbers within a matrix template. This approach offers distinct advantages, as it circumvents challenges associated with unclear handwriting and enables the unambiguous identification of individual students, as multiple students may share the same name but must possess a unique identification number.

## III. Dataset

The dataset utilized for training the neural network to recognize student ID numbers comprises scanned portions of exams containing completed matrix templates. This dataset was collected from a course conducted at the Faculty of Electrical Engineering and Computing, University of Zagreb.

The matrix template represents a table of size $n \times 10$, where $n$ corresponds to the number of digits in the student's identification number. Specifically, the student personal number at the aforementioned university consists of 10 digits, resulting in a matrix representation of size $10 \times 10$ (see Fig. 1).

The data set comprises a total of 1703 examples, consisting of images featuring labeled matrix templates and their associated labels. Out of these, 1658 examples contain correctly filled matrix templates, while the remaining examples encompass incorrectly filled matrix templates.

A correctly filled matrix template (CFMT) contains precisely one blackened row in each column, representing the corresponding digit. Conversely, incorrectly filled matrix templates include instances where multiple blackened fields are present in the same column, no blackened fields are present in any of the columns, or the matrix template is visibly filled incorrectly (e.g., crossed out).

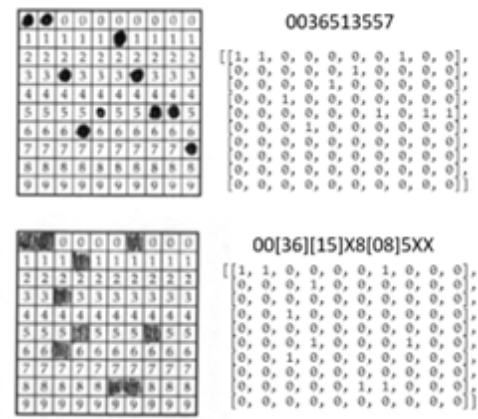

Fig. 1. An exemplary empty matrix template.

The labels are represented by a 2D vector of numbers, wherein the blackened positions in the table are denoted with the number 1 within the vector, while all other positions are marked with the number 0. For the sake of clarity, the vector can also be presented as a textual record. Columns with a single unit in the column are displayed with the respective digit, empty columns are indicated with the symbol "X", and in the case of two or more units in a column, all marked digits are enclosed within square brackets (e.g., "[34]") (see Fig. 2).

Fig. 2. An example of a correctly and incorrectly filled matrix template with associated labels and textual records.

## IV. Architecture

Recognizing student ID numbers poses a unique challenge that cannot be simply categorized as a classic image classification problem. To address this specific problem, this article presents a specially designed architecture optimized for solving this task. The architecture employed in this research to recognize student personal numbers is based on a modified version of the U-Net [4] architecture. The modifications primarily focus on the decoder part of the architecture.

Traditionally, the decoder part of the architecture consists of layers that gradually increase the resolution from the lowest level to match the input layer resolution. However, in this model, the decoder part extends only up to the layer that increases the resolution to the size $16 \times 16$. Following the U-Net segment of the architecture, a convolution layer (transformation layer) is introduced, with a kernel size adjusted to produce an output layer of the desired resolution, specifically $10 \times 10$. Finally, a convolutional output layer with a sigmoid activation function is utilized to generate prediction labels (see Fig. 3).

The decoder part consists of resolution reduction blocks. Each block comprises a convolutional layer with a kernel size of 5, a LeakyReLU activation function (with alpha set to 0.2), and a dropout layer with a dropout rate of 0.1.

The upscaling block in the architecture incorporates a convolutional transpose layer with a kernel size of 3 and a stride of 2. This layer facilitates the expansion of resolution from the previous layer. The resulting layer is then concatenated with the corresponding layer from the downsampling part of the network, followed by dropout and another convolutional layer.

The number of channels for each layer in the model is determined by the hyperparameter "channels," which is specified as a list with four values. The first value in the list corresponds to the number of channels for layers 0 and 1 in the encoder. The second value is utilized for layers 2 and 3, the third value for layers 4 and 5, and the fourth value for the bottleneck layer. The decoder part employs the same number of channels as the corresponding encoder layers but in the reverse order. For instance, if the number of channels in layers 0 and 1 is set to 16, then the decoder layers connected to those encoder layers will also have 16 channels. The transformation layer of the model has a fixed number of channels determined by the "last_channel" parameter.

The two aforementioned parameters provide the flexibil-ity which enables achieving an optimal balance between network size and prediction quality. By adjusting the number of channels in the layers, the network's capacity and its ability to learn relevant features can be controlled.

This custom architecture combines various layers and techniques to attain high accuracy and reliability in recognizing student ID numbers.

## V. Training

The network was trained using the following procedures. Initially, the images were resized to a target size of 128x128 pixels to ensure uniformity in the input grid size. Prior to being fed into the network, the images underwent random augmentation to enhance the training data. Augmentations applied include rotation (rt), shearing (sh), and scaling (sc). The degree of augmentation is controlled by the hyperparameter $p_{org}$, which determines the expected percentage of non-augmented examples in each batch. Each augmentation type has an augmentation factor ($\gamma_{rt}$, $\gamma_{sh}$, $\gamma_{sc}$) between 0 and 1, and based on $p_{org}$, a parameter $\mu$ satisfying the equation "(1)" is obtained using the golden-section search. Finally, each augmentation probability ($p_{rt}$, $p_{sh}$, $p_{sc}$) is calculated by multiplying augmentation factor by the calculated $\mu$. The augmentation factors are set as $\gamma_{rt} = 0.4$, $\gamma_{sh} = 0.3$, and $\gamma_{sc} = 0.3$. The value of the hyperparameter $p_{org}$ is set to 0.5, indicating that approximately 50% of the images in each batch are expected to be original. This approach strikes a balance between using original images for the model to learn the inherent patterns and utilizing augmented examples to enable the model to capture variations and enhance its robustness. The Adam optimizer is employed for network training, while binary cross-entropy serves as the loss function. The training process consists of 150 epochs.

Additionally, the initial learning rate is set to 0.0015. After every 20 epochs, the learning rate decreases by 10%, resulting in a new learning rate that is 90% of the previous value. This learning rate schedule helps the model to converge effectively and avoid overshooting the optimal solution.

$$p_{org} = (1 - \mu\gamma_{rt})(1 - \mu\gamma_{sh})(1 - \mu\gamma_{sc}) \qquad (1)$$
$$p_{rt} = \mu\gamma_{rt} \qquad (2)$$
$$p_{sh} = \mu\gamma_{sh} \qquad (3)$$
$$p_{sc} = \mu\gamma_{sc} \qquad (4)$$

## VI. Results

This section presents the results of detecting student personal numbers using the custom architecture proposed in this article. Additionally, a comparison of model performance with varying complexities is presented.

### A. Evaluation Metrics

Given the unique nature of the problem, the evaluation metrics are tailored to suit the specific requirements of the task. Let $y^{(i)}$ represent the 2D vector of the actual label for the $i$-th example, and let $\hat{y}^{(i)}$ represent the 2D vector of the predicted label for the same example. We define the equality of these two vectors if all their corresponding elements are equal, formally expressed as:

$$y^{(i)}_{ij} = \hat{y}^{(i)}_{ij} \quad \forall i, j$$

The accuracy (ACC) of the prediction is defined as the mean value of the indicator function $I(y^{(i)}, \hat{y}^{(i)})$ (Eq. 5), where $I$ is 1 if the vectors are equal, and 0 if they are not, for all examples.

$$ACC = \frac{1}{N} \sum_{i=1}^{N} I(y^{(i)}, \hat{y}^{(i)}) \qquad (5)$$

$$I(y^{(i)}, \hat{y}^{(i)}) = \begin{cases} 1, & \text{if } y^{(i)} = \hat{y}^{(i)} \\ 0, & \text{otherwise} \end{cases} \qquad (6)$$

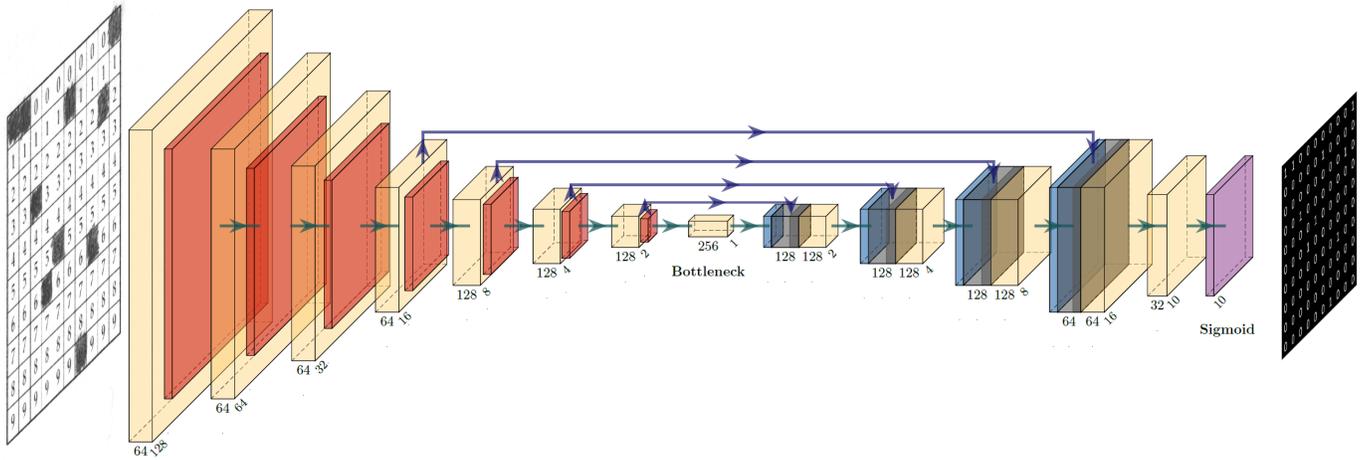

Fig. 3. Network architecture.

The alpha error (critical error) represents the model's error in which $y^{(i)} \neq \hat{y}^{(i)}$ and $\hat{y}^{(i)}$ satisfies CFMT. This error is considered critical because tags that satisfy CFMT do not require interaction with a rectifier, making it challenging to detect. The alpha error rate represents the proportion of alpha error examples in relation to the total number of predictions that satisfy CFMT.

On the other hand, the beta error represents the model's error in which $y^{(i)} \neq \hat{y}^{(i)}$ and $y^{(i)}$ satisfies CFMT. Although less critical, the beta error still requires manual review, even though the matrix record is correctly filled. The beta error rate represents the proportion of beta error examples in relation to the total number of tags that satisfy CFMT.

### B. Model Performance Evaluation

To evaluate the model, the k-fold method was employed to divide the dataset into 5 parts. Each model was trained 5 times, with one part used as a validation dataset and the remaining 4 parts used for training. The reported metrics in Table I are the mean values obtained from the 5 training runs. K-fold validation improves the robustness of performance estimation by mitigating the impact of random validation set selection, particularly when dealing with limited or variable datasets, reducing potential bias resulting from a single split of the data.

The results show that all three models achieve remarkably high accuracy and low alpha and beta error rates. As expected, the model with the highest number of parameters demonstrates the best performance across all metrics. The experimental findings demonstrate that the best model achieves an accuracy of 97.02%. Additionally, with an alpha error rate of 0.43% and a beta error rate of 1.63%, the model exhibits high reliability and suitability for real-life scenarios. These results highlight the practical applicability of the proposed model in accurately predicting the correct student identification number. While the most complex model demonstrates the best performance, it is important to consider alternative models when deploying on devices with lower hardware capabilities, prioritizing the optimal balance between speed and accuracy. Therefore, the selection of the model should carefully account for the hardware limitations of the target device, ensuring efficient and effective deployment.

### VII. Conclusion and Future Work

This paper introduces an architecture for recognizing student personal numbers from a matrix template. The proposed model, based on a modified U-Net architecture, achieves high accuracy in detecting student identification numbers for automated student identification systems. Additionally, the paper outlines the annotation method for marked examples and presents evaluation metrics for assessing model performance. The experimental results demonstrate the accurate detection of matrix template entries, with the best model achieving an accuracy of 97.02% on validation data, with an alpha error of 0.43%, and a beta error of 1.63%. Based on these metrics, it can be concluded that the model successfully detects student numbers from the matrix template. In future work, there is potential to further enhance the model's performance. One avenue for exploration is expanding the dataset to encompass diverse scenarios and variations of student ID numbers. By incorporating a larger and more comprehensive dataset, the model can improve its generalization capabilities.

TABLE I
Model Performance Comparison

| Channels | Last Channels | Parameters | Epoch Time | ACC | α Error Rate | β Error Rate |
|---|---|---|---|---|---|---|
| 16, 16, 32, 64 | 8 | 349,473 | 6 | 0.9497 | 0.0062 | 0.0344 |
| 32, 32, 64, 128 | 16 | 1,396,033 | 7 | 0.9667 | 0.0049 | 0.0181 |
| 64, 64, 128, 256 | 32 | 5,580,417 | 8 | 0.9702 | 0.0043 | 0.0163 |